\documentclass[english, letterpaper, 10pt]{article}

\usepackage[T1]{fontenc}
\usepackage{textcomp}

\usepackage{graphicx}
\usepackage{amssymb,amsmath}
\usepackage{fixltx2e}
\usepackage{fancyhdr}
\usepackage{url}

\usepackage[numbers,sort&compress]{natbib}

\usepackage{dsfont}
\usepackage[top=1in, bottom=1.25in, left=0.875in, right=0.875in]{geometry}
\usepackage{bm}

\usepackage{mathptmx}   
\usepackage{titlesec}
\usepackage{titling}

\usepackage{booktabs}
\usepackage{graphicx}
\usepackage{multirow}
\usepackage{subcaption}

\usepackage[english]{babel}
\usepackage[utf8]{inputenc}
\usepackage{amsmath}
\usepackage{amsfonts}
\usepackage{graphicx}
\usepackage{enumitem}
\usepackage[colorinlistoftodos]{todonotes}
\usepackage{abstract}

\usepackage{indentfirst}

\usepackage{times}

\usepackage{titlesec}
\titleformat{\section}{\bfseries\centering\fontsize{11pt}{13pt}\selectfont}{\thesection.}{4pt}{\uppercase}
\titleformat{\subsection}{\bfseries\fontsize{11pt}{13pt}\selectfont}{\thesubsection}{6pt}{}
\titlespacing*{\section}{0pt}{10pt plus 8pt}{4pt}
\titlespacing*{\subsection}{0pt}{6pt}{3pt}

\usepackage{titling}
\pretitle{\vspace{-15mm}\begin{center}\bfseries\LARGE}
\posttitle{\end{center}}
\preauthor{\begin{center}\large}
\postauthor{\end{center}\vspace{-15mm}}

\usepackage{etoolbox}
\apptocmd{\thebibliography}{\setlength{\itemsep}{-2pt}}{}{}

\newlength{\myitemsep}
\setlist[itemize]{itemsep=-1pt, topsep=2pt}
\setlist[enumerate]{itemsep=-1pt, topsep=1pt}

\title{Influence of Resampling on Accuracy of Imbalanced Classification}
\author{E. Burnaev, P. Erofeev, A. Papanov
\\
Institute for Information Transmission Problems (Kharkevich Institute) RAS
}
\date{}

\begin{document}
\pagenumbering{gobble}
\maketitle

\renewcommand{\abstractname}{\vspace{0pt}\fontsize{11pt}{13pt}\selectfont \uppercase{Abstract}\vspace{-4pt}}
\begin{abstract}
\normalsize
In many real-world binary classification tasks (e.g. detection of certain objects from images), an available dataset is imbalanced, i.e., it has much less representatives of a one class  (a \emph{minor class}), than of another.
Generally, accurate prediction of the minor class is crucial but it's hard to achieve since there is not much information about the minor class.
One approach to deal with this problem is to preliminarily \emph{resample} the dataset, i.e., add new elements to the dataset or remove existing ones.
Resampling can be done in various ways which raises the problem of choosing the most appropriate one.
In this paper we experimentally investigate impact of resampling on classification accuracy, compare resampling methods and highlight key points and difficulties of resampling.
\vspace{6pt}
\\
\textbf{Keywords:} binary classification, class imbalance, resampling, Bootstrap oversampling, RUS, SMOTE

\end{abstract}

\section{Introduction}

In this paper, we focus on binary classification tasks with imbalanced datasets,
i.e., two-class datasets having much less representatives of one class (\emph{minor~class}) than of another (\emph{major~class}).
Many real-world data analysis problems have inherent peculiarities which lead to unavoidable imbalances in available datasets.
Examples of such problems include detection of whether a patient is ``cancerous'' or ``healthy'' from mammography image~\cite{cancer}, oil spill detection from satellite images~\cite{oilspill}, network intrusion detection~\cite{net-intrusion}, detection of fraudulent transactions on credit cards~\cite{cardfraud}, diagnosis of rare diseases~\cite{med}, prediction and localization of failures in technical systems~\cite{faultloc, pm-itas}, etc.
Indeed, target events (frauds, failures, diseases, etc.) in these problems are rare, so there is much less information about them than about non-events (normal transactions, correct behavior, etc.).
Hence an attempt to formulate any of these problems as a binary classification task leads to the class imbalance in the available dataset. Note that this effect is unavoidable since it is caused by the nature of the problem.

Moreover, the minor class is often the one of prime interest~\cite{learn-imbalanced}.
E.g., in the examples above it corresponds to target events whose accurate prediction is crucial for applications.
However, standard classification models (e.g., logistic regression, SVM, decision trees, nearest neighbors) treat all classes as equally important and thus tend to be biased towards the major class in imbalanced problems~\cite{svm-unbalanced, logreg-unbalanced, dtree-unbalanced-2}.
This may lead to poor prediction of minor class elements with the average quality of prediction being good.
For example, consider a process with target events occurring just $1$\% of all cases.
If a classification model always gives a `no-event' answer it is wrong just $1$\% of all cases which seems to be a good quality on average.
But such model is absolutely useless for the minor class prediction.
Thus imbalanced classification problems require a special treatment.

There are several ways~\cite{learn-imbalanced} to increase importance of the minor class.
\begin{itemize}
\item
Adapt a probability threshold for classifiers which yield probabilities of belonging to classes.

\item
Modify a loss function, e.g., assign more weight to misclassification of minor class elements compared to misclassification of the major class.

\item
Resample a dataset in order to soften or remove class imbalance. Resampling may include:
	\begin{itemize}
	\item[$\diamond$]
    \emph{oversampling}, i.e. addition of synthesized elements to the minor class;

    \item[$\diamond$]
    \emph{undersampling}, i.e. deletion of particular elements from the major class;

    \item[$\diamond$]
    combined resampling, i.e. both oversampling and undersampling.
	\end{itemize}
\end{itemize}
Below we focus on the latter approach.
It is convenient and widely used since it allows to tackle imbalanced tasks using standard classification techniques.
On the other hand, this approach comprises problem of resampling method and resampling amount (i.e., how many observations to add or drop) selection.

In this paper, we consider three widely used resampling methods (see section~\ref{res-methods}) and two simplest strategies of resampling amount selection (see section~\ref{sec:mult-select-str}).
We experimentally explore their influence on quality of classification of more than $1000$ imbalanced datasets.
The exploration shows that resampling is capable of improving the quality for most datasets, however, resampling method and amount have to be selected properly, see section \ref{results} for more detailed discussion.

\section{Notations and Problem Statement}
\label{notation}

In this section, we introduce notations which will be used further. Consider dataset of $\ell$~elements with binary labels: $S = (X_i, y_i)_{i = 1}^{\ell}$, where $X_i \in \mathbb{R}^d$, $y_i \in \{0, 1\}$.
Denote $C_0(S) = \{(X_i,y_i)\in S \mid y_i=0\}$ and $C_1(S) = \{(X_i,y_i)\in S \mid y_i = 1\}$.
Let label $0$ corresponds to the major class, label $1$ corresponds to the minor class, then $|C_0(S)| > |C_1(S)|$.
To measure a degree of a class imbalance for a dataset, we introduce an \emph{imbalance ratio}
$I\!R(S) = \frac{|C_0(S)|}{|C_1(S)|}$.
Note that $I\!R(S) \geq 1$ and the higher it is, the stronger imbalance of $S$ is.

The final goal is to learn a classifier using imbalanced training sample~$S$.
This is done in two steps.
First, the dataset $S$ is \emph{resampled} using a resampling method $r$: some observations in $S$ are dropped or some new synthetic observations are added to $S$.
The result of resampling is a dataset $r(S)$ with $I\!R(r(S)) < I\!R(S)$.
Next, some standard classification model $h$ is learned on $r(S)$, which gives classifier
$h_{r(S)}: \mathbb{R}^d \rightarrow~\{0, 1\}$ as a result.

Performance of a classifier is determined by a predefined \emph{classifier quality metrics} $Q(h_{S_{train}},S_{test})$ which takes as input classifier $h_{S_{train}}$, testing dataset~$S_{test}$ and yields higher value for better classification.
In order to determine performance of $r$ and $h$ on the whole dataset~$S$ regarding metrics~$Q$, we use a standard procedure based on $k$-fold cross-validation~\cite{stat-learning} which yields $Q^{CV}(S)$.

\section{Overview of Resampling Methods}
\label{res-methods}

Every resampling method $r$, considered in this article, works according to the following scheme.

\begin{enumerate}
\item
Takes input:
\begin{itemize}
\item
  dataset $S$ as described in section \ref{notation};

  \item
  \emph{resampling multiplier} $m>1$ which determines resulting imbalance ratio as $I\!R(r(S)) = \frac{1}{m} \cdot I\!R(S)$ and thereby controls amount of resampling;

  \item
  additional parameters, which are specific for every particular method.
\end{itemize}

\item
Modifies given dataset by adding synthesized objects to the minor class (\emph{oversampling}), or by dropping objects from the major class (\emph{undersampling}), or both. Details depend on the method used.

\item
Outputs
resampled dataset $r(S)$ with $d$ features and imbalance ratio $I\!R(r(S)) = m \cdot I\!R(S)$.

\end{enumerate}

In this paper, we consider the most widely used resampling methods: Random Oversampling, Random Undersampling and Synthetic Minority Oversampling Technique (SMOTE).

\subsection{Random Oversampling}

Random oversampling~\cite{learn-imbalanced} (ROS, also known as bootstrap oversampling) takes no additional input parameters.
It adds to the minor class new $(m-1)|C_1(S)|$ objects.
Each of them is drawn from uniform distribution on $C_1(S)$.

\subsection{Random Undersampling}
Random Undersampling~\cite{learn-imbalanced} (RUS) is an undersampling method, it takes no additional parameters.
It chooses random subset of $C_0(S)$ with $|C_0(S)|\frac{m-1}{m}$ elements and drops it from the dataset.
All subsets of $C_0(S)$ have equal probabilities to be chosen.

\subsection{SMOTE}

Synthetic Minority Oversampling Technique (SMOTE)~\cite{smote} is an oversampling method, it takes one additional integer parameter $k$ (number of neighbors).
It adds to the minor class new synthesized objects, which are constructed in the following way.
\begin{enumerate}
\item
Initialize set as empty: $S_{new} := \emptyset$

\item
Repeat the following steps $(m-1)|C_1(S)|$ times:
\begin{enumerate}[label=(\roman*)]
\item
Select one random element $x_i \in C_1(S)$.

\item
Find $k$ minor class elements which are nearest neighbors of $x_i$.
Randomly select one of them (call it $x_j$).

\item
Select random point $x$ on the segment connecting $x_i$ and $x_j$.

\item
Assign minor class label to the newly generated element $x$ and store it: $S_{new} := S_{new} \cup \{(x, 1)\}$.
\end{enumerate}

\item
Add generated objects to the dataset: $\tilde{S} = S\cup S_{new}$.
\end{enumerate}

\subsection{Other Resampling Methods}

There are several other resampling methods which are less widely used:
Tomek Link deletion~\cite{one-sided-selection}, One-Sided Selection~\cite{one-sided-selection}, Evolutionary Undersampling~\cite{eus}, borderline-SMOTE~\cite{borderline-smote}, Neighborhood Cleaning Rule~\cite{ncl}.
There exist also procedures combining resampling and classification in boosting:
SMOTEBoost~\cite{smoteboost}, RUSBoost~\cite{rusboost}, EUSBoost~\cite{eusboost}.
These methods are not examined in this paper because we aimed to explore only the most widespread resampling methods.

\section{Methodology of Comparison}

\subsection{Data, Classifiers, Quality Evaluation}
\label{sec:data-clf-q}
We used two pools of datasets for experimental comparison:
artificial pool with $\sim\! 1000$ datasets and real pool \cite{data-1, data-2} with $\sim\! 100$ datasets.
Artificial datasets were drawn from a Gaussian mixture distribution.
Each of two classes is represented as a Gaussian mixture with not more than $3$ components.
Number of features varies from $6$ to $40$, size of dataset from $200$ to $1000$, $I\!R$ from $0.05$ to $0.35$.
Real-world datasets have come from different areas: biology, medicine, engineering, sociology.
All features are numeric or binary, their number varies from $3$ to $1000$.
Size of dataset varies from $200$ to $1000$, $I\!R$ from $0.02$ to $0.75$.

We ran learning process as described in section~\ref{notation}, for each dataset we varied classification model, resampling method and resampling multiplier.
Bootstrap, RUS and SMOTE with $k=5$ (as taken in \cite{smote}) were considered as resampling methods.
We varied resampling multiplier from $1.25$ to $10.0$.
As classification model $h$ we used Decision Trees, $k$-Nearest Neighbors, and Logistic Regression with $\ell_1$ regularization.
Optimal parameters of classification models were selected by cross-validation.

Area under precision-recall curve $Q_{PRC}$ was used as a quality metric.
To evaluate quality of resampling and classification, we performed $10$-fold cross-validation and calculated $Q_{PRC}^{CV}$ --- average of $Q_{PRC}$.
Results of experiments are described by values of $Q_{PRC}^{CV}$ for each dataset, resampling method, resampling multiplier and classification model.

\subsection{Resampling Multiplier Selection}
\label{sec:mult-select-str}

We considered two strategies of resampling multiplier selection:
\begin{itemize}
\item
equalizing strategy, \emph{EqS}: select multiplier providing balanced classes ($I\!R=1$) in resulting dataset;

\item
CV-search, \emph{CVS}: select optimal multiplier (i.e., providing maximum of $Q^{CV}$) by cross-validation.

\end{itemize}
The equalizing strategy seems to be reasonable as it removes class imbalance which we initially tried to tackle.
It is quick and widely used.
CV-search may provide better quality but it is more time-consuming.

\subsection{Dolan-More Curves}

To compare resampling methods, we use Dolan-More curves~\cite{dolan-more} which are built in the following way. Let $\{r_1, \ldots, r_n\}$ be the set of considered resampling methods, $\{S_1, \ldots S_T\}$ be the set of tasks (datasets),
$q_{ti}$ be the quality of the method $i$ on the dataset $t$.
For each method $i$ we introduce $p_i(\beta)$, a fraction of datasets, on which the method $i$ is worse than the best one not more than $\beta$ times:
$$
p_i(\beta) = \frac{1}{T} \left| \left\{t: q_{ti} \geq \frac{1}{\beta} \max\limits_{i} q_{ti} \right\}\right|, \ \beta \geq 1.
$$
For example, $p_i(1)$ is a fraction of datasets where the method~$i$ is the best.

A graph of $p_i(\beta)$ is called Dolan-More curve for the method $i$.
This definition implies that the higher the curve, the better the method.
Note that Dolan-More curve for a particular method depends on other methods considered in comparison.

\section{Results}
\label{results}

\begin{figure}[t!]
\center{
\includegraphics[resolution=900, scale=0.98]{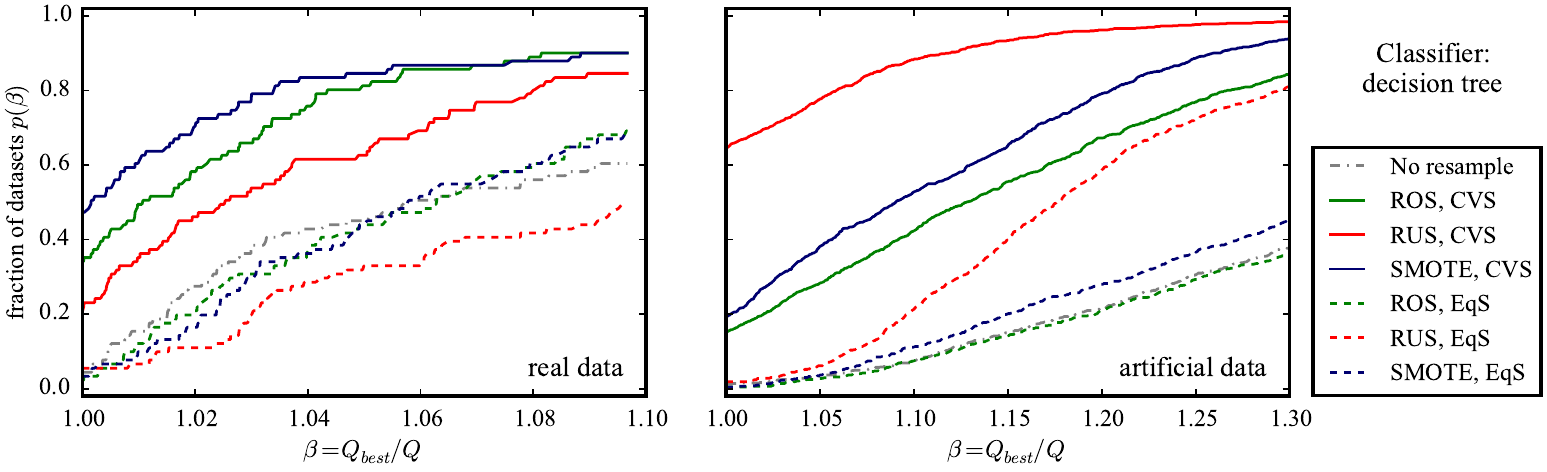}
\\
\vspace{-1mm}

\includegraphics[resolution=900, scale=0.98]{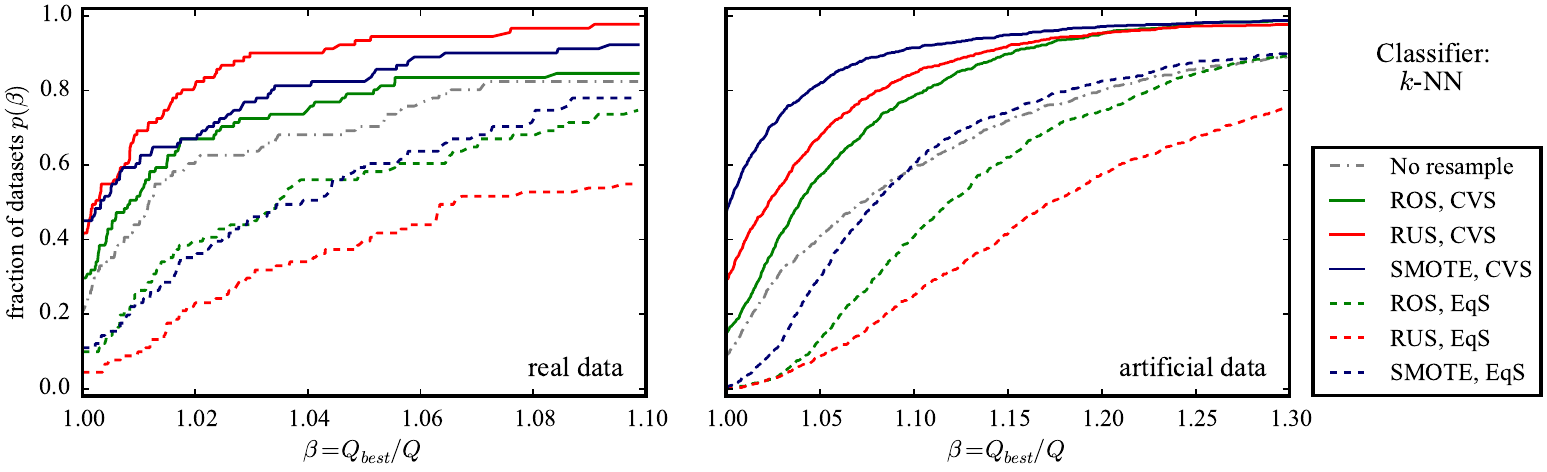}
\\
\vspace{-1mm}

\includegraphics[resolution=900, scale=0.98]{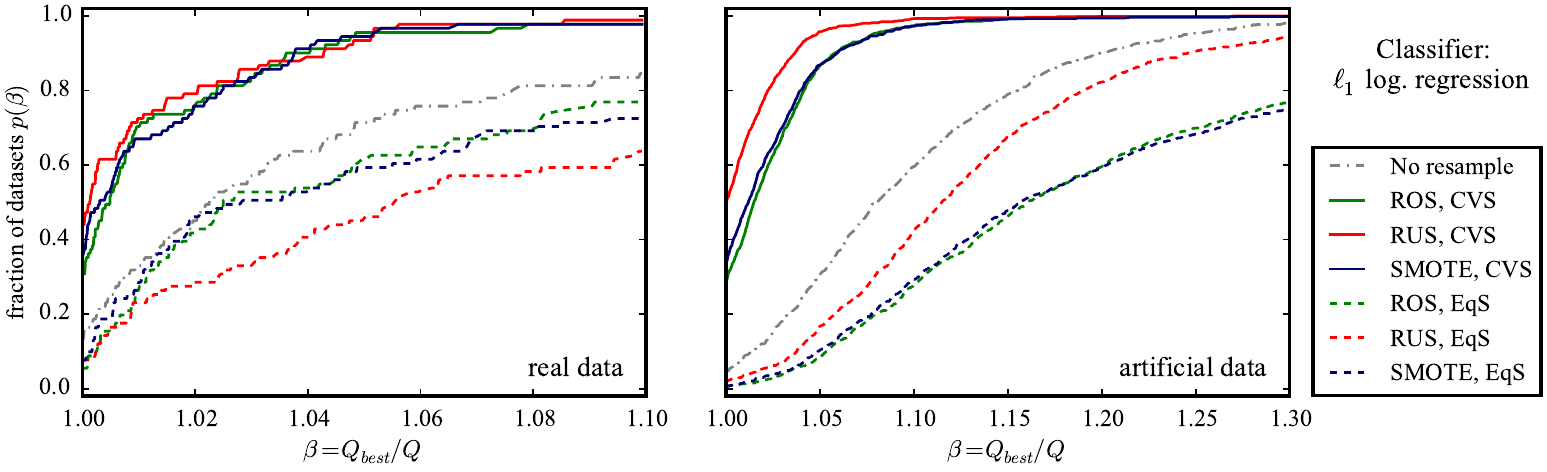}
\\
\vspace{-1mm}
}
\caption{Dolan-More curves for metric $Q_{PRC}^{CV}$}
\label{pic:dmc}
\end{figure}

Figure \ref{pic:dmc} provides Dolan-More curves for quality of classification with: no~resampling, ROS, SMOTE ($k$=5) and RUS.
Here we consider both multiplier selection strategies (see section \ref{sec:mult-select-str}).
In order to compare resampling methods, the curves are plotted separately for each classification model and also for real and artificial data.

\begin{figure}[t!]
\center{
\includegraphics[resolution=900, scale=1]{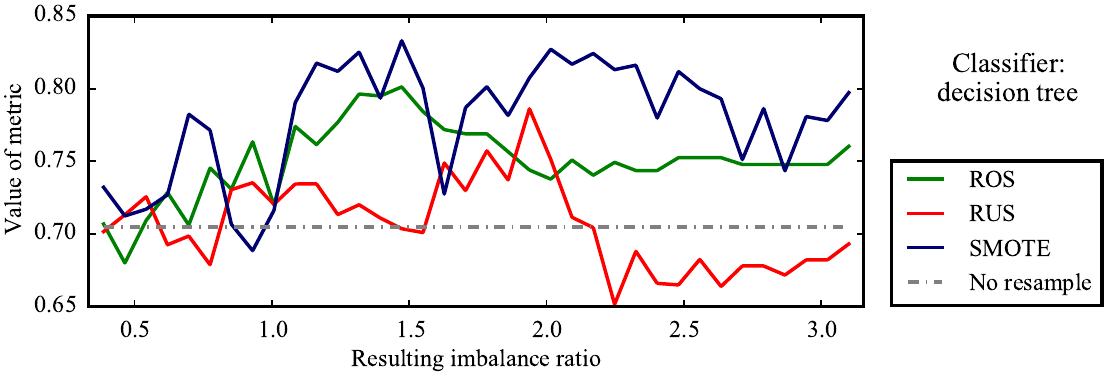}
}
\caption{Value of $Q_{PRC}^{CV}$ vs. resulting $I\!R$
for dataset ``Delft pump 1x3'' from \cite{data-2}}
\label{pic:q-vs-ir}
\end{figure}

First of all, these curves show that influence of resampling on the quality strongly depends on resampling multiplier.
That is, all resampling methods with CV-search of multiplier improve the quality on most datasets, especially for Decision trees and Logistic regression.
In contrast, the equalizing strategy of multiplier selection (EqS) shows much lower quality, and it is even worse than no resampling for $k$-Nearest neighbors and Logistic regression.
Such low performance of EqS is not surprising because dependence of quality on the multiplier may be sophisticated:
see for example figure \ref{pic:q-vs-ir}, which represents this dependence for a certain real dataset.

Secondly, performance of resampling method depends on the classifier used, and there is no method that would always outperform the others.
For example, RUS with CVS strategy is best when used with $k$-Nearest neighbors on real datasets, but with Decision tree on the same datasets it performs worse than other methods with CVS.

Thirdly, impact of resampling on quality depends on the data it is applied to.
RUS EqS used with Decision tree demonstrates this distinctly:
it is worse than no~resampling for the real datasets but outperforms it on the artificial data.
Since artificial datasets are quite similar (see section~\ref{sec:data-clf-q}),
this means that they all have similar characteristics that result in higher quality for Decision tree with RUS EqS.
At the same time, real datasets are more diverse and most of them have another characteristics that result in lower quality.

Finally, classification without resampling is the best choice in some cases.
E.g., for Logistic regression it is about $15$\% of real datasets and $5$\% of artificial.
Therefore not all imbalanced datasets have to be resampled to achieve better classification quality.

The overall conclusion is the following. 
Resampling improves classification of imbalanced datasets in most cases if a method and a multiplier are selected properly.
But if not, resampling may have negative effect on quality of classification.
Thus if one decides to resample imbalanced dataset, one has to select a method and a multiplier in order to get an actual quality improvement.
Moreover, there is no universally good choice of how to resample the dataset.
That is, the best resampling method and multiplier for one dataset can be worse than no~resampling for another.
E.g., equalizing classes with resampling doesn't always improve the quality.
Unexpectedly, but in some cases the best choice of resampling is not to resample at all.
So, to improve quality of classification, one has to determine optimal resampling method (also considering no~resampling) and multiplier in every particular imbalanced task.

\textbf{Acknowledgement}: The research was conducted in IITP RAS and supported solely by the Russian Science Foundation grant (project 14-50-00150).

\renewcommand\refname{\uppercase{References}}

\end{document}